\documentclass{article}

\usepackage{arxiv}

\usepackage{cite}
\usepackage{amsmath,amssymb,amsfonts}
\usepackage{algorithmic}
\usepackage{graphicx}
\usepackage{textcomp}
\usepackage{epstopdf}
\usepackage[T1]{fontenc}
\usepackage{lmodern}
\usepackage{multirow}
\usepackage{csvsimple}

\usepackage{verbatim} 
\usepackage{soul} 
\usepackage{wasysym} 

\usepackage[T1]{fontenc}
\usepackage[utf8]{inputenc}
\usepackage[emoticons]{emoji}

\usepackage{graphics}
\usepackage[bookmarks=false]{hyperref}
\usepackage{ifpdf} \ifpdf
\usepackage{longtable}

\usepackage{booktabs}
\usepackage{subcaption}

\title{Using GAN-based models to sentimental analysis on imbalanced datasets in education domain}

\author{
 Ru Yang, Maryam Edalati\\
 Department of Computer Science\\
	Norwegian University of Science and Technology\\
	Gj\o vik, Norway  \\
	\texttt{ruya@stud.ntnu.no, maryame@stud.ntnu.no} \\
	}

\usepackage{blindtext}
\usepackage{graphicx}

\begin{document}
\maketitle

\begin{abstract}

While the whole world is still struggling with the COVID-19 pandemic, online learning and home office become more common. Many schools transfer their courses teaching to the online classroom. Therefore, it is significant to mine the students' feedback and opinions from their reviews towards studies so that both schools and teachers can know where they need to improve. This paper trains machine learning and deep learning models using both balanced and imbalanced datasets for sentiment classification. Two SOTA category-aware text generation GAN models: CatGAN and SentiGAN, are utilized to synthesize text used to balance the highly imbalanced dataset. Results on three datasets with different imbalance degree from distinct domains show that when using generated text to balance the dataset, the F1-score of machine learning and deep learning model on sentiment classification increases $2.79\% \sim  9.28\%$. Also, the results indicate that the average growth degree for CR100k is higher than CR23k, the average growth degree for deep learning is more increased than machine learning algorithms, and the average growth degree for more complex deep learning models is more increased than simpler deep learning models in experiments.
\end{abstract}


\section{Introduction}
\label{sec:Introduction}
As a result of the COVID-19 pandemic, many schools and universities have pivoted from traditional, in-person physical classes towards online courses. Technological development in recent years has led to improvements to the technology behind online courses, and as a result, more students in developing countries and remote areas can take courses from top universities through their computer or mobile phone. This presents an opportunity for potentially reducing the education inequality across the globe~\cite{lee2001}. There are already multiple platforms providing free online courses, such as Massive Open Online Courses (MOOC), Coursera, Khan Academy, Udemy and edX, providing lectures on a wide variety of subjects \cite{imran2012multimedia,moore2011learning}.

Many online learners do not primarily rely on online courses to complete their courses, but to enhance the effect of conventional learning techniques, meeting new classmates or reviewing specific topics \cite{davis2017follow}. As such, it is not appropriate to use completion rates as an indicator to evaluate the effectiveness of MOOCs \cite{dalipi2017analysis} for learners with other needs than completing their course and earning a certificate \cite{hew2020predicts}. The authors in \cite{chuang2016harvardx} make the point that the effectiveness of MOOCs could be mischaracterized if completion rates are overemphasized. This also includes drop-out rates \cite{dalipi2018mooc,imran2019predicting} and other issues relating to the MOOCs~\cite{dalipi2016towards}. 

Therefore, learning institutions examine the feedback from students regarding their experiences with online learning platforms so that both professors and the platforms themselves can learn which aspects need to be altered and improved. For example, in \cite{Kastrati2020} the paper indicates that feedback from students helped the platform to implement a co-creation process during the projects life cycle. Additionally, professors can benefit from the feedback from students to help them understand student behaviour and refine the contents of the courses \cite{rowe2017feelings,rowe2018understanding}.

Student feedback is usually structured to include not only close-ended questions but also open questions allowing students to express their thoughts about various aspects of teaching \cite{Kastrati2020AspectBasedOM}. It is essential to examine students' sentiments about specific aspects of this feedback, as seeking out the opinions of others is a prevalent practice when it comes to a decision making \cite{cambria2016affective}.

Often the amount of data from these student reviews will be so large that it would be impractical to process these reviews manually. There is also the potential for language barriers to complicate the task, for example, with terms and abbreviations used by student-aged people on the web \cite{estrada2020opinion}. To analyze this sort of textual feedback accurately would require state-of-the-art technology, such as traditional machine learning or deep neural networks. Additionally, this task comes at a time when the practice of opinion mining is increasingly controversial. Kastrati et al. provided a detailed systematic mapping study on sentiment analysis of student's feedback \cite{kastrati2021sentiment}.

The development of deep learning technology has made enormous contributions to Natural Language Processing \cite{kastrati2019impact}. For deep learning models, there are several implementations of Neural Networks of Deep learning, such as the Convolutional Neutral Networks (CNN), Recursive Neutral Networks (RNN), Long Short-Term Memory Networks (LSTM), BERT, which efficiently carry out the task of opinion mining. These deep learning models have been widely employed for opinion mining in various domains, including movies \cite{dos2014deep}, social media platforms i.e. Twitter, Facebook \cite{imran2020cross,batra2021evaluating,Kastrati:2021LowResource}, e-commerce \cite{vanaja2018aspect}, eLearning \cite{Kastrati2020}, tourism \cite{afzaal2019tourism}, to name just a few. However, these deep learning models will usually need a large amount of training data to achieve competitive classification performance. Additionally, deep learning models usually have to pass through preprocessing steps as the text they work with are not structured for direct input of neural models. The preprocessing steps usually include "tokenizing" and "normalizing". Tokenizing is to split sentences into smaller elements called tokens by using the common delimiters and then to convert the sentence attributes into a set of numeric attributes representing word occurrence information \cite{lwin2020feedback}. Normalizing is to replace words that have a similar meaning with a single word – for example, could words like "studied", "studies", "studying" be replaced with the word "study". Stemming and lemmatization are two commonly used normalization techniques.
On the other hand, there are multiple machine learning algorithms such as Bernoulli Naive Bayes, Support Vector Machine (SVM), LinearSCV, Random Forest. Most of these algorithms can be found in the Python library scikit-learn. Before the data is fed into the neural network or machine learning algorithm, it is necessary to perform the preprocessing step to improve the algorithms' performance. Real-world statistics are often strident, incomplete and incompatible. It is important to preprocess and concentrate the data before the dataset can be used for machine learning \cite{katragadda2020performance}.

\begin{figure}[ht]
    \centering
    \includegraphics[width=0.6\textwidth]{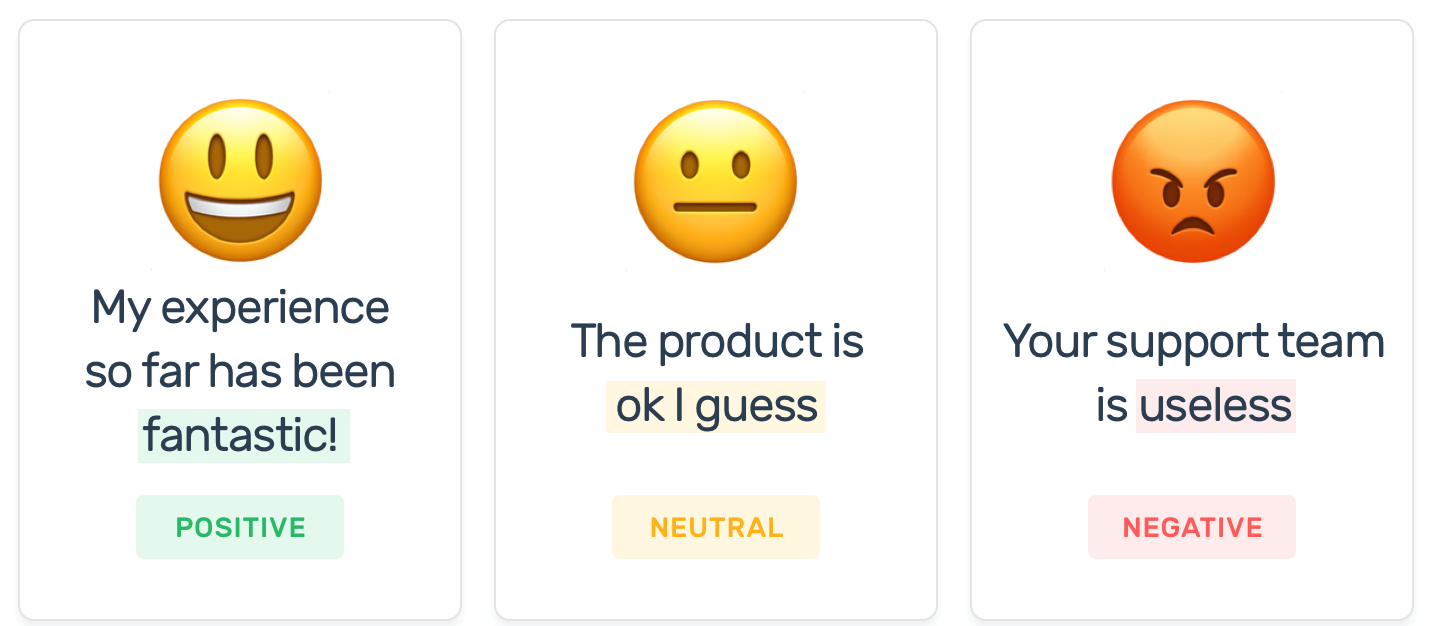}
    \caption{3-Classes Sentiment Analysis\cite{wolff-2020}}
    \label{fig:3-classes}
\end{figure}

Sentiment analysis using both machine learning and deep learning algorithms is commonly used to classify a text to its respective class. Based on varying data and reasons behind it, sentiment analysis could be binary (positive or negative) or multi-class problem that has been shown in Figure \ref{fig:3-classes} (3 or more classes). Natural language processing has developed a lot in the past ten years, benefiting from the rapid development of graphics chip, which has much powerful computing capability compared with the past.
{}

However, in the classification task, data imbalance is a common issue that would affect the performance of models adversely, and the imbalanced data problem is more often in the educational domain than other domains \cite{yang2020sentiment}. This means that one sentimental label is far more than others because people tend to give positive reviews towards teachers or courses. But the minority class are often critical in revealing issues of courses. Besides, in the most application field, acquiring the same amount of sample for each class is almost impossible in real life. In general, the commonly seen samples are far more than uncommonly seen samples. Researchers often address this problem using the data sampling technique and generating synthetic data from original training samples. Deep learning methods with the generative adversarial network (GAN) can achieve good performance on image tasks. But different from general image, synthetic text often comes across context and semantic information lost.  
Linguistic communication is a process of encoding by the speaker and decoding by the listener. Depends on the people involved in conversation, decoding and encoding of information can be different. The sentence used in communication is often composed of compressed information ignoring the details for the statistical model. This limits the model generating sentence with good structure leading the language model walks to the wrong way in generating text when meeting metaphor or hidden words relationship. For example, "the stormy ocean was a raging bull" is an example of a metaphor. Based on the background information, both speaker and listener can understand that this sentence describes "the stormy ocean is quite dangerous", while many language models may infer that Ocean is a bull and generate text-based on this context. 

Text generation model can be evaluated by human or linguistic experts or by using text generation metrics to evaluate performance objectively. The commonly used text evaluation metric is BLEU (Bilingual Evaluation Understudy), which is context evaluation. BLEU tries to evaluate the generated text quality by comparing the similarity between generated text and real text. The closer synthetic text is to real text, the higher the performance. In addition to text quality, we also need to pay attention to the diversity of generated text. $NLL_{gen}$ and $NLL_{gen}$ proposed in \cite{liu2020catgan} evaluate the diversity of the generated text.

This project aims to train the STOA text generation GAN models, select and utilize the GAN model with better performance to generate negative and neutral reviews to balance the original highly imbalanced dataset and evaluate how different sentiment classification model with different dataset respond. The overall problem statement can be described as below:

\emph{Analyzing the impact of synthetic text generation on sentiment classification task of the highly imbalanced dataset using deep learning and machine learning.}

\section{Related Work}
\label{ch:bg}




In the past ten years, with the rapid development of computation capability of graphics, and more educational resources \cite{kastrati2020wet} and frameworks \cite{kastrati2019integrating}, there are more and more sentiment analysis research based on deep learning and traditional machine learning algorithms \cite{daudpotapotential}. Also, GAN network that were initially used on computer vision for synthesizing pictures starts to be used in text generation.

Sindhu el. \cite{sindhu2019aspect} built a learning model with two LSTM layers to analyze the sentiment polarization of students' reviews. The two layers works as different classifiers. One layer is used for aspect extraction while another is for classifying the sentiment of the reviews as positive, negative or neutral. The aspects from the output of the first layer would be used as the input of the second layer. The public dataset comprising of restaurant reviews is used to test the model trained by students' reviews from reviews of physical classroom courses of the university. The results of the research indicate that the two-LSTM-layer model used in this paper gets 91\% accuracy in extracting aspect and 93\% in classifying sentiment of the students' reviews. For the public restaurant reviews, the model achieves 82\% accuracy in extracting aspects and 85\% accuracy in sentiment classification.
\begin{figure}[ht]
    \centering
    \includegraphics[width=0.8\textwidth]{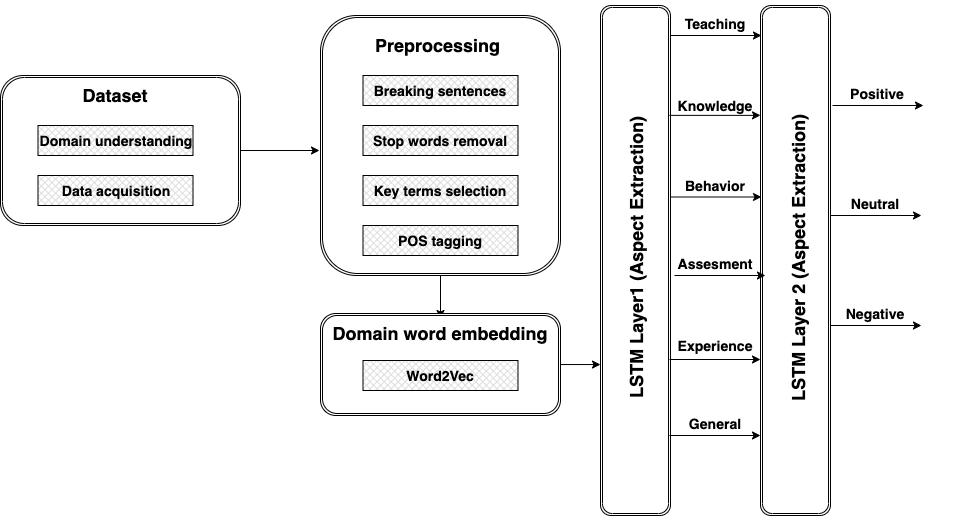}
    \caption{Flow diagram of methodology \cite{sindhu2019aspect}}
    \label{fig:two layers lstm}
\end{figure}{}

The scientific paper \cite{Kastrati2020AspectBasedOM} compared the conventional machine learning and deep learning algorithms on sentiment analysis based on $21940$ students' reviews from e-learning platform. During their experiments, they used 4 traditional machine learning algorithms: SVM, Naive Bayes, Boosting, Decision Tree, and built one 1D-CNN deep learning model to extract aspect and analyze sentiment. Their research results show that 1D-CNN achieved better performance on sentiment analysis getting 88.2\% on F1 scores but traditional machine learning models are better at aspect extraction.

Anna el. \cite{koufakou2016using}  conducted a survey containing several free text questions and got 204 answers from the survey. They classified the 204 students' feedback into 162 positive reviews and 42 negative reviews. In their experiment, the traditional machine learning algorithms Naive bayes and $K$-nearest algorithms to predict the students' reviews are positive or negative. Besides, cosine similarity is utilized to measure the similarity and leave-one-out cross in order to validate the result. The authors compared the results from their research with the Recursive Neural Tensor Network (RNTN) \cite{socher2013recursive} method. The paper \cite{koufakou2016using} indicates that RNTN \cite{socher2013recursive} has better performance in Precision but worse Recall and Accuracy.

\begin{table}[ht]
\centering
\begin{tabular}{l|l}
\hline
\textbf{Type}                     & \textbf{Model} \\ \hline
\multirow{5}{*}{Deep Learning}    & LSTM\_3b       \\
                                  & CNN\_5a        \\
                                  & CNN\_10a       \\
                                  & CNN\_LSTM\_7a  \\
                                  & BERT           \\ \hline
\multirow{8}{*}{Machine Learning} & Multinomial NB \\
                                  & KNN            \\
                                  & Decision Tree  \\
                                  & B4MSA          \\
                                  & Bernoulli NB   \\
                                  & SVC            \\
                                  & Linear SVC     \\
                                  & Random Forest  \\ \hline
Evolutionary Algorithm            & EvoMSA         \\ \hline
\end{tabular}
\caption{14 algorithms used in the experiment of \cite{estrada2020opinion}}
\label{table:1}
\end{table}
Katragadda et al. \cite{katragadda2020performance} researched sentiment analysis using several supervised machine learning algorithms and one deep learning model in order to classify the feedback as positive, negative or neutral. Their dataset includes thirty thousand feedback containing anonymous personal information, reviews, and students' emotion. Also, the dataset is classified into two categories: linear dataset with same properties, and non-linear dataset. The article shows that machine learning models get better results on linear dataset. More specifically, Naive Bayes model gets 50\% accuracy after the precision and recall are calculated inside of it. Beside, SVM algorithms achieves 60.8\% accuracy and the deep learning model gets 88.2\% accuracy much beyond the conventional machine learning algorithms. 

In the research job \cite{elia2019assessing} the big data mining framework were built to achieve the instant monitoring of the students' satisfaction on online learning platforms. The framework contains both Data Management and Data Analytic techniques aiming to fill the gap between limited focus of big data on educational fields and rapid development of big data techniques. The framework is made to be able to connect to various kinds of data sources easily. The discussion of the classes in forum and stuents' reviews are used as 2 main data sources used in this research. The significant part of the their framework is called Analysis Engine which is used to analyze sentiment, cluster, and classify the reviews.During the pre-processing steps, the text features are collected using the below TF-IDF \cite{na2004effectiveness} equation:
\begin{equation}
TF*log(\frac{N}{DF})
\end{equation}
where TF is the number of keywords occurrences in the current processing file, N is the counting of keywords occurrences in all files, and DF is the counting of total files in the experiment. The dataset contains 15000 balanced textual reviews. Then the balanced dataset is fed into two machine learning models Linear SVM in order to train the machine learning models, and Cross Validation is used to validate the results. The statistical supervised learning algorithm (SVM) that "one-against-one" strategy has been used with a "max wins" selection strategy. When clustering the dataset, TF-DF algorithm is implemented by the authors to transform textual reviews to number firstly. Then they the K-means models to extract clusters for survey form and feedback collected from the forum. Finally, the controlled experiment involving few e-learning students and a small test data is conducted showing the functionalities of the framework and its potential value. Besides, the authors point out the future direction of the research. On the one hand, the connection between sentiment of the lessons and students' final mark could be built. On the other hand, other information, e.g. login information, admitted lessons, as well as contents posted on social media could be incorporated.

The research work in \cite{hew2020predicts} investigated the factors influencing students' MOOC satisfaction and give us an extended general understanding of those factors. In their research, students' satisfaction is regarded as an crucial metric defining the success of MOOC. They classify the independent variables into learner-level and course-level variables and utilize those two aspects variables to predict students satisfaction. How those independent variables of student and course level affect the dependent variable, -- MOOC students satisfaction , is evaluated by the authors. The dataset is downloaded from a public course website \textit{Class Central} where people are able to download class metadata and feedback regarding the respective course. They collected feedback from 6391 students from this website. During the experiment, several traditional machine learning models, e.g k-nearest neighbors regression \cite{altman1992introduction}, gradient boosting trees \cite{friedman2001greedy}, support vector machines \cite{cortes1995support}, logistic regression \cite{cox1958regression}, and naive Bayesian \cite{murphy2012machine}, are used to classify the emotion polarization. The algorithm achieving the best performance among all models was then chosen to predict aspect labels for the left part of unlabeled  reviews. Their research results show that the gradient boosting tree got best results among all traditional machine learning models. As for the computation of sentiment polarity scores of the input text, the TextBlob is utilized that is a public free text processing software. The output scores for each review from TextBlob ranges from -1.0 to 1.0. The authors identified three crucial factors having statistically strong associations with learner satisfaction regarding learner sentiment in the conclusion part, which are content, assessment, as well as instructor. But there are no directive connections between course structure, video, and interactions and MOOC students' satisfaction \cite{hew2020predicts}. There are two disadvantages of this sentiment analysis research. On the one hand, eighty percent of the feedback from the \textit{Class Central}  is written by those students who finished the whole courses. On the other hand, because of the intrinsic difference of the data the real randomness between the dependent variable -- MOOC satisfaction,  and those independent variables was not addressed.

Lwin et al. \cite{lwin2020feedback} conducted the research on not only open text reviews but also on rating scores. The dataset is gotten from the online survey form from the students of the university, which all the questions are just rating value questions except the last question that is free text based and used to collect feedback on classes and teachers. In textual comments analysis, the textual feedback were classified into two categories: negative, and positive, while rating scores were classified into five types: Worse, Bad, Neutral, Good and Excellent. The labeling work of the dataset utilizes the K-means clustering algorithms to pre-label the huge amount of the feedback data. Then the labeled dataset is fed into multiple conventional machine learning models. The six algorithms, -- Logistic Regression, Multiplayer Perceptron, Simple Logistic Regression, Support Vector Machine, LMT and Ransom Forest, are selected to make comparisons in terms of performances. The situation is different for the sentiment analysis of textual comments. Firstly, people label each sentence as positive or negative manually, and conduct pre-processing steps on those reviews using the open-source library NLTK in order to analyze textual comments. The authors conclude that SVM gets the best results on rating score classification and Naive Bayers algorithms yields best performance for textual comment analysis.

The research work \cite{estrada2020opinion} conducted the experiments using 8 conventional machine learning models, 5 deep learning models and one evolutionary model.These fourteen algorithms they have used in their research have been presented in the Table \ref{table:1}. Two different kinds of dataset that is crawled from the HTML code of YouTube and other online learning platforms, is used: eduSERE and SentiTEXT. eduSERE is able to represent learning-center sentiments like engaged, excited, disappointed, and bored while the other is only with two polarities: $positive$ and $negative$. The authors then build a sentimental dictionary connecting between words in text and emotion of the reviews. An algorithm based on the word count to classify the sentiment and the learning-centered emotion is proposed in order to pre-label the feedback they have collected. There are some reviews which were hard to classify and so were removed by the authors when checking the pre-label results. Finally, they choose the accuracy as the metrics to evaluate the performance of the models and algorithms. Their research presents that BERT and EvoMSA get better performance with 93\% accuracy on SentiTEXT classification and descent accuracy of 84\% and 83\% on EduSERE classification. In the last, the integration between the models and an intelligent learning environment is performed by them. The intelligent learning environment is developed using Java. On the last part of the article, the authors concluded that genetic EvoMSA model have the best performance after adopting more knowledge and being optimized by macro-F1 aiming to solve the unbalanced dataset problem.

\begin{figure*}[ht]
    \centering
    \includegraphics[width=\textwidth]{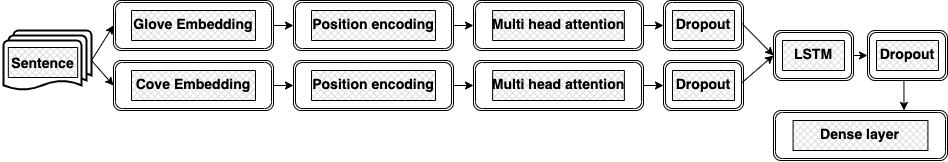}
    \caption{Multi-head attention model architecture}
    \label{fig:1}
\end{figure*}{}

The researchers came up with a fusion deep learning model in \cite{sangeetha2020sentiment} to analyze learners' reviews. The data set is the Vietnamese Student Feedback Corpus (UIT-VSFC) \cite{van2018uit}, which contains 16,000 reviews from Vietnamese students, which are then machine-translated into English for later use. The fusion model contains of a multi-head layer and an LSTM layer, and its structure is shown in the Figure \ref{fig:1}. At the beginning, the feedback is fed into two different training embedding: Glove embedding and Cove embedding. Through embedding, the word information in the sentence would be changed to position information. Then use multiple attention blocks to calculate the weighted sum of multiple attentions rather than only concerning a single attention. The attention mechanism is used to assign weights to context words to find the words that determine the sentiment of the input sentence. Then the researchers utilize different dropout rates in order to avoid over-fitting problem, as well as combine the outputs of two different embedding models (1024 features) as the input of LSTM. Finally, through dropout and intensive layer, one of three emotions is classified: positive, negative, and neutral. The results show that the performance of the proposed multi-head attention fusion model is better than the single LSTM model, LSTM model + attention model and multi-head attention model.

The paper \cite{gao2019target} displays a novel target-dependent sentiment classification with BERT. The target-dependent Bert(TD-BERT) \cite{gao2019target} takes the positioned output at the target words rather than the first [CLS] tag because one sentence can refer to many targets with their own context. Then TD-BERT is followed by a max-pooling operation before the output is fed to the next fully-connected layer. The proposed model can extracts multiple targets from one sentence and then predicts the sentiment polarity of the sentence by combining the sentiment polarity of each target. The results \cite{gao2019target} indicates combining TD-BERT with complicated neural network does not show much value, sometimes even having worse performance than vanilla BERT-FC (fully connected). Besides, the accuracy is improved when the target information is incorporated. Most models often do not pay much attention to semantics \cite{kastrati2019impact,imran2012semantic}. BERT and all other deep learning models are only good at NLP tasks, not much when it comes to natural language understanding\footnote{https://medium.com/ontologik/time-to-put-an-end-to-bertology-or-ml-dl-is-not-even-relevant-to-nlu-e5ba6fc53403}. Such issues can be addressed employing ontologies, better vector space representation models \cite{kastrati2019performance,kastrati2014adaptive}, and objective and semantic metrics \cite{kastrati2016semcon,kastrati2015semcon}.

The authors in the paper \cite{estrada2020opinion} compared the performance of fourteen models in sentimental analysis. The models that were used include one evolutionary algorithm, eight Machine Learning models, and five Deep Learning models, and have been shown in \ref{table:1}. For the dataset, they created two corpus: SentiTEXT and eduSERE. SentiText has $positive$ and $negative$ polarity while the later one can represent learning-center emotions like engaged, excited, bored, and frustrated. The dataset is obtained by crawling the HTML code of YouTube and other educational platforms. Then a emotional dictionary building connections between words in text and sentiment polarity is created. The dictionary is combined with a simple algorithm based on word count to determine the polarity and the learning-centered emotion to pre-label the collected opinions. During this process, the opinions that were difficult to classify was removed when the computer experts reviewed the pre-label results. Besides, Accuracy is chosen to evaluate the performance of the models. The result shows that BERT and EvoMSA achieve the best accuracy with 93\% on SentiTEXT classification and close accuracy with respective 84\% and 83\% on EduSERE classification. Finally, the authors integrate the models into an intelligent learning environment developed by Java. In the conclusion, they pointed out that EvoMSA based on Genetic Programming achieved best results by incorporating extra knowledge and using macro-F1 optimization to solve unbalanced datasets problem.

\begin{figure}[ht]
    \centering
    \includegraphics[width=0.8\textwidth]{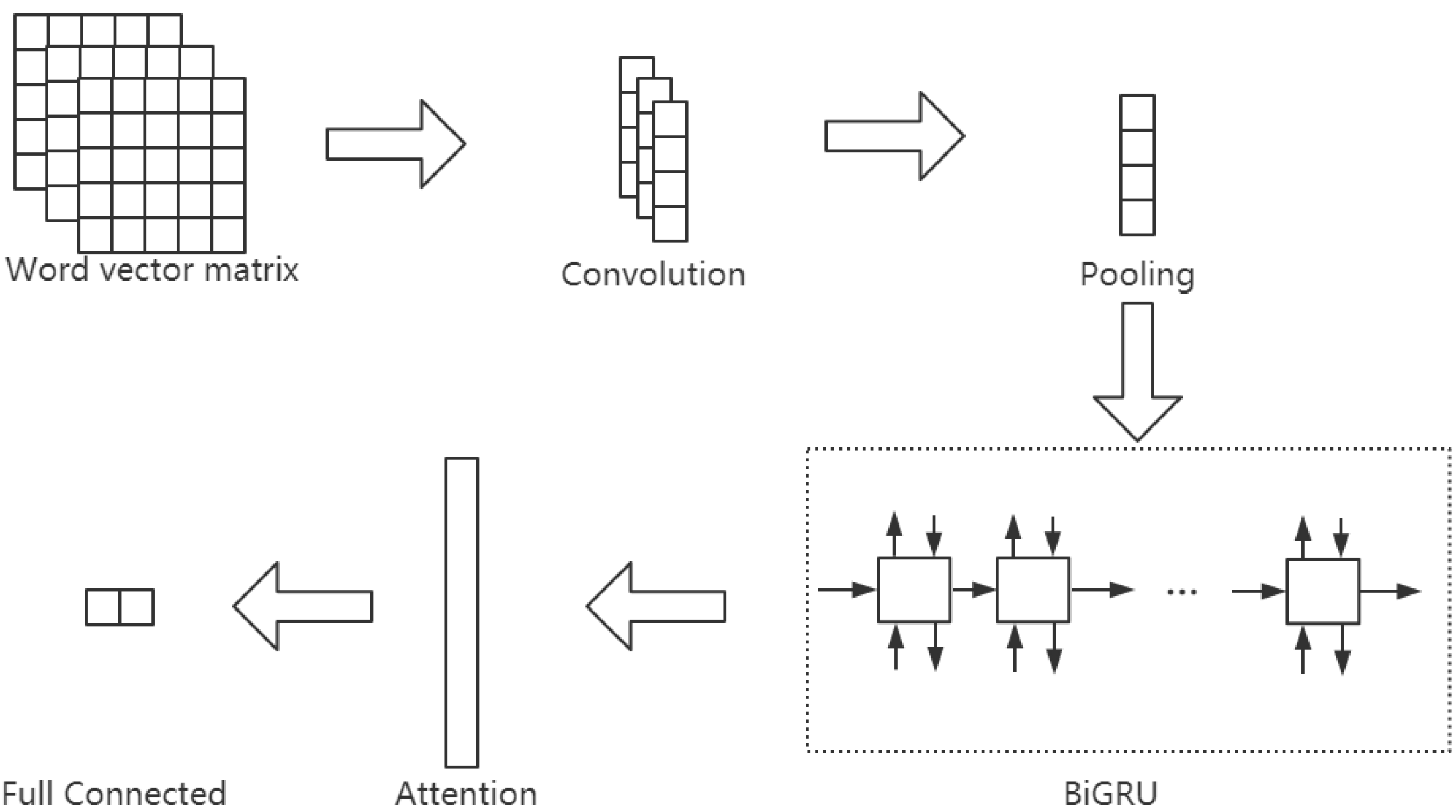}
    \caption{Flow diagram of methodology \cite{sindhu2019aspect}}
    \label{fig:SLCABG model}
\end{figure}

The paper \cite{yang2020sentiment} proposes a novel sentiment analysis model-SLCBG combining Convolutional Neural Network (CNN) and attention-based Bidirectional Gated Recurrent Unit (BiGRU) which has been shown in Figure \ref{fig:SLCABG model}. Their model is based on sentiment lexicon that is used to improve the sentiment features in the reviews. The dataset is e-commerce product reviews collected from a Chinese online shopping website. Finally they compared their model with other sentiment analysis models proving their model overcome disadvantages of other deep learning models and performance comparison results have been shown in Table \ref{tab:peformance comparison}.

\begin{table}[ht]
\centering
\begin{tabular}{|l|l|l|l|l|}
\hline
Model           & Accuracy & Precision & Recall & F1     \\ \hline
NaiveBayes \cite{dey2016sentiment}      & 57.9\%   & 55.6\%   & 79.2\% & 65.3\% \\ \hline
SVM \cite{Kastrati2020AspectBasedOM}            & 67.7\%   & 93.8\%   & 38.4\% & 54.5\% \\ \hline
CNN \cite{lakshmi2017sentiment}             & 90.9\%   & 91\%     & 90.2\% & 90.6\% \\ \hline
CNN+Attention \cite{shin2016lexicon}   & 91.4\%   & 90.8\%   & 91.6\% & 91.2\% \\ \hline
BiGRU \cite{chen2019gated}          & 92.6\%   & 91.1\%   & 94.1\% & 92.6\% \\ \hline
BiGRU+Attention \cite{zhou2019improved} & 93.1\%   & 92.8\%   & 93.2\% & 93\%   \\ \hline
SLCBG \cite{yang2020sentiment}          & 93.5\%   & 93\%     & 93.6\% & 93.3\% \\ \hline
\end{tabular}
\caption{Performance comparison of different sentiment analysis models.\cite{yang2020sentiment}}
\label{tab:peformance comparison}
\end{table}

{}

In the paper \cite{shaikh2021towards}, the authors did a relevant project as this project. For the highly data imbalanced problem, they solved it by using text sequence generation algorithms to help text classification on highly imbalanced dataset in specific field. For text generation models, they used the newly proposed GPT-2 and LSTM based text generation model to balance the highly unbalanced text dataset. In their experiments, three highly imbalanced datasets from different domains were used and the results show that the performance of the same deep learning network model improves 17\%.

The authors in \cite{yang2020sentiment} presented results of systematic mapping study of sentiment analysis using NLP, machine learning and deep learning within educational field. They used PRISMA framework to instruct the searching process which includes papers published from 2015 to 2020 in electric research database. 92 relevant study from 612 research are found that are gotten from results of sentiment analysis of students' feedback on online learning platform. Their results show that sentiment analysis is still in fast development stage especially in deep learning application though there are challenges. The authors also indicate that structured dataset, standard solution as well as sentiment expression and detection need more attention.

\section{Methods}
\label{chap:methods}



This section describes the models and algorithms used in our experiment along with initial text processing steps. 

\subsection{Text preprocessing}
Following text processing steps are applied to clean the dataset. 

\begin{itemize}
    \item Tokenization
    \item Lower casing
    \item Stop words removing
    \item Normalization
    \item Lemmatization
\end{itemize}

In addition to this, we further applied language detection. One of the reviews dataset is scraped from the the online learning platform, Coursera. This online learning platform not only has English courses but also contains many courses of other languages. Because our project focuses on sentiment analysis rather than multilingual classification, language detection and extraction of English reviews are necessary. 

For the language detection, the open source library Polyglot\footnote{https://github.com/aboSamoor/polyglot} is used, which is able to detect 165 languages. Polyglot library relies on pycld2 library. And it is dependent on cld2 library for language detection of text. Sometimes one review would contain more than one language. If this happens, the detector can give us the most likely languages in the text and their confidence level respectively.


\subsection{Text generation models}

In this article, we employed two text generation models explained in following subsections:

\begingroup
\setlength{\LTright}{\LTleft}
\begin{table}[ht]
\begin{tabular}{p{1cm}|l|p{2cm}|l|p{8cm}}
\toprule
\textbf{Type} & \multicolumn{1}{c|}{\textbf{Name}} & \multicolumn{1}{c|}{\textbf{Author}} & \multicolumn{1}{c|}{\textbf{Year}} & \multicolumn{1}{c}{\textbf{Article Name}} \\ \midrule
\multirow{8}{*}{General} & SeqGAN & Yu et al. & 2017 & SeqGAN: Sequence Generative Adversarial Nets with Policy Gradient\cite{yu2017seqgan} \\ \cmidrule(l){2-5} 
 & LeakGAN & Guo et al. & 2018 & Long Text Generation via Adversarial Training with Leaked Information\cite{guo2018long} \\ \cmidrule(l){2-5} 
 & MaliGAN & Che et al. & 2017 & Maximum-Likelihood Augmented Discrete Generative Adversarial Networks\cite{che2017maximum} \\ \cmidrule(l){2-5} 
 & JSDGAN & Li et al. & 2019 & Adversarial Discrete Sequence Generation without Explicit Neural Networks as Discriminators\cite{li2019adversarial} \\ \cmidrule(l){2-5} 
 & RelGAN & Nie et al. & 2018 & RelGAN: Relational Generative Adversarial Networks for Text Generation\cite{nie2018relgan} \\ \cmidrule(l){2-5} 
 & DPGAN & Xu et al. & 2018 & DP-GAN: Diversity-Promoting Generative Adversarial Network for Generating Informative and Diversified Text\cite{xu2018dp} \\ \cmidrule(l){2-5} 
 & DGSAN & Montahaei et al. & 2021 & DGSAN: Discrete Generative Self-Adversarial Network\cite{montahaei2021dgsan} \\ \cmidrule(l){2-5} 
 & CoT & Lu et al. & 2019 & CoT: Cooperative Training for Generative Modeling of Discrete Data\cite{lu2019cot}\\ \midrule
\multicolumn{1}{p{1.5cm}|}{\multirow{2}{*}{Category}} & SentiGAN & Wang et al. & 2018 & SentiGAN: Generating Sentimental Texts via Mixture Adversarial Networks\cite{wang2018sentigan}\\ \cmidrule(l){2-5} 
\multicolumn{1}{l|}{} & CatGAN & Liu et al. & 2020 & CatGAN: Category-aware Generative Adversarial Networks with Hierarchical Evolutionary Learning for Category Text Generation\cite{liu2020catgan} \\ \bottomrule
\end{tabular}
\caption{10 GAN text generation model}
\label{tab:gan}
\end{table}
\endgroup

As we can see from the Table \ref{tab:gan}, there are two state of the art GAN model used to generate text of different category. So, these two GAN models are selected in our experiments to generate text. Next, we will describe these two category GAN models in the below two sections.

\subsubsection{SentiGAN}

The GAN used in generating text usually have low quality, lack of diversity and mode crashing problem. In the paper \cite{wang2018sentigan}, the authors proposed a new framework SentiGAN. This framework contains multiple generator and a classifier used to distinguish real and generated text. In their framework, multiple generators are trained at the same time aiming to generate text of different category without any supervision. The aim based on penalty are proposed in their generator to force the generators to generate text with diversity. Besides, each generator can focus on generating text of a specific category without needing to worry about the other categories. 

SentiGAN is composed of multiple generators based on LSTM and a classifier. They are trained at the same time. Similar to the paper \cite{yu2017sequence}, the authors regard sequence generation process as sequence decision process. They also apply randomly initialization strategy to parameters of each generator model and Monte Carlo search is used to search the appropriate behavior value. Then they use classifier to evaluate the generated text which is then used to instruct generators leaning. What is different from the previous models, their model contains multiple generators and one classifier. Firstly, a new aim based on penalty is proposed. This aim adopts more appropriate measures aiming to minimize the the overall penalty in biggest degree rather than maximizing rewards value. 

The authors think target based on penalty can force each generator synthesize text with specific sentimental polarity label instead of generating safe and good samples repeatedly. Also, different generators are separate from each other and could focus on generating text of its own category without influence from other types of sentiments. They think this would improve the sentimental accuracy. Besides of sentimental polarity, other metrics such as fluency, novelty, diversity, and intelligibility are tested by using classifier with good performance to evaluate the generated text. The adversarial training process of SentiGAN is from \cite{wang2018sentigan}.

\begin{figure}[ht]
    \centering
    \includegraphics[width=0.8\textwidth]{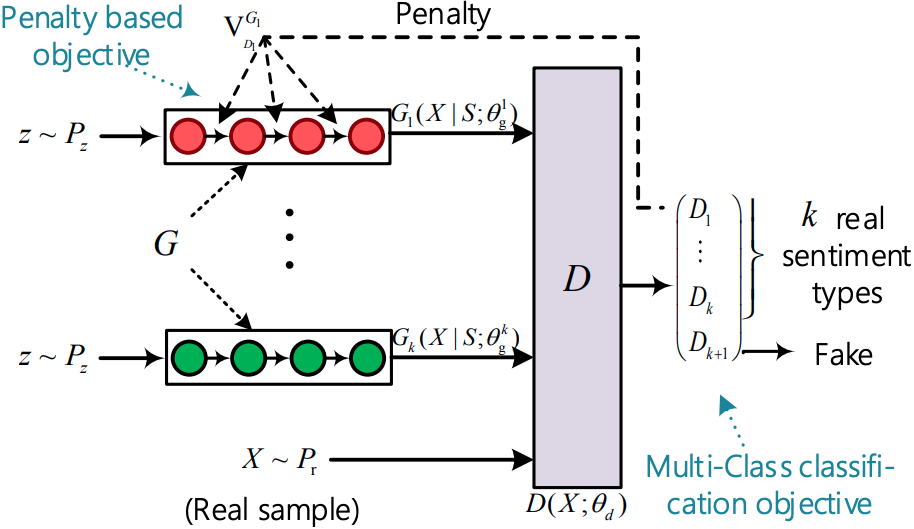}
    \caption{SentiGAN with mixture adversarial networks\cite{wang2018sentigan}}
    \label{fig:sentigan}
\end{figure}

The model structure of SentiGAN has been shown in Figure \ref{fig:sentigan}. If we assume that we are generating text with multiple sentimental polarities, $k$ generators $\left \{ G_i(X\mid (S;\theta _g^i)) \right \}_{i = 1}^{i = k}$ and one discriminator $D(X;\theta_d)$ would be used, where $\theta _g^i$ and $\theta_d$ are the parameters of the $i$-th generator and classifier. $\theta _g^i$ and $\theta_d$ would use initialize the input of generator by using pre-checked input noise sampling from normal distribution. The whole framework consists of two adversarial learning aims: generator learning and classifier learning. The aim of $i$-th classifier is to generate text of the $i$-th sentimental label. They hope the generated text could deceive the classifier. In other words, it aims at minimizing the objective based on penalty. At the same time, the aim of classifier is to distinguish fake text from the real samples as much as possible. These are the aims of multi classes classification they have adopted.

\subsubsection{CatGAN}
Because generative adversarial network could achieve competitive results in text generation, it has been adopted in generating text of different classes in research work \cite{wang2018sentigan}, which is known for sentiGAN. But according to the study \cite{liu2020catgan}, the complicated model structure and learning strategy limit the performance of GAN and increase the instability of training process. So they proposed a category aware adversarial network. 

\begin{figure}[ht]
    \centering
    \includegraphics[width=\textwidth]{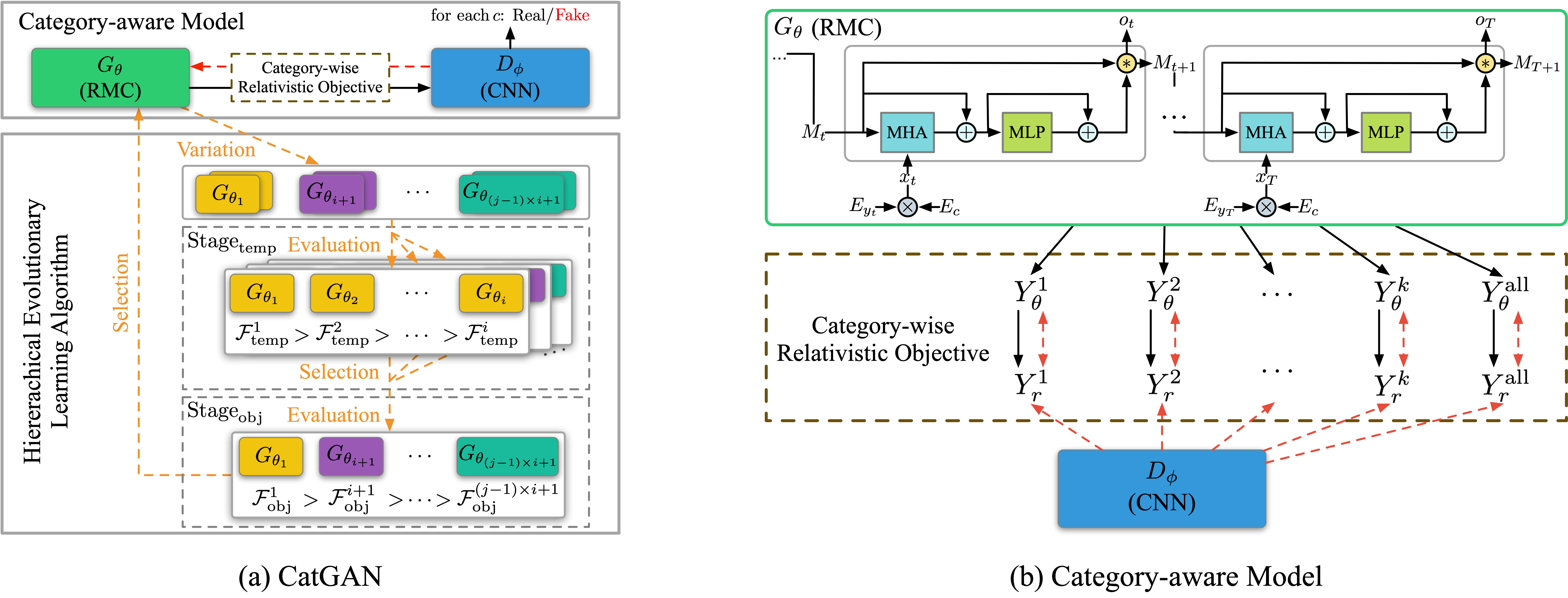}
    \caption{Category-aware GAN with hierarchical evolutionary learning \cite{liu2020catgan}}
    \label{fig:catgan}
\end{figure}

The network structure has been shown in \ref{fig:catgan}. As we can see, the model is composed of a category aware model and hierarchical evolutionary algorithm that is used to train model. The category model measures the difference between real samples and generated samples on each category and try to use the aim of minimizing the difference to instruct model to generate text of specific category with high quality. The generator is based on relational memory core to generate text with a specific category. At the same time, there is classifier trying to distinguish real samples and generated samples of each category. In their model, gradients can be transmitted to generator from classifier directly with the help of Gumbel Softmax function. In order to train the model and improve performance, they also proposed a hierarchical evolutionary algorithm. The evolutionary algorithm aims to stabilize training process and get balanced between quality and diversity when training CatGAN. According to the study, if we only focus on samples quality, mode crashing and over-fitting problem would inevitably happen. This is the reason why the authors introduce hierarchical evolutionary algorithm. It evolves the group of parents generator $G_{\theta}$ by all kinds of mutation strategies under given environment $D_{\phi}$. It allows model to keep the offspring who has better performance with high diversity and high quality.
The Figure \ref{fig:catgan} shows the model structure of the category-aware GAN model.

\subsection{Sentiment classifiers}

In this study, we employed both conventional (Decision Tree, AdaBoosting, SVM, Naive Bayes) and deep learning models (RNN, BiLSTM, CNN, GRU). The systematic mapping study \cite{yang2020sentiment} on sentiment classification of students' reviews with deep learning indicates that the most frequently algorithms used in sentiment analysis are SVM and Naive Bayes that are part of supervised learning algorithms. In addition to SVM and Naive Bayes, decision tree, k-NN and Neural Network algorithms are also often used in sentiment analysis. We further compared and analysed the results using accuracy, precision, recall, and F1 score as evaluation metrics. The results are discussed in the following section.

\section{Datasets}
\label{sec:datasets}
Three datasets of different size and from different domains are used in the experiments. Two datasets are from the education domain, and another is from the entertainment including game and movie reviews. 

\subsection{CR23k}
The first dataset contains 21937 course reviews from the online learning platform Coursera, which is obtained from the paper \cite{Kastrati2020AspectBasedOM}. The samples of this dataset have been shown in \ref{tab:crsample}, in which each review is labelled with course information and sentimental polarity. Hereafter, we would use CR23k to refer to this dataset for clearness and convenience.
\begin{table}[hbp!]
\centering
\begin{tabular}{@{}llp{0.85\textwidth}@{}}
\toprule
\textbf{Course} & \textbf{Label} & \textbf{Reviews} \\ \midrule
C & P & end of course project was challenging and fun. lots of opportunity to learn how to debug memory issues with valgrind. \\\hline
C & NEU & teaches you how to use gdb and debug code effectively. challenging and engaging homework. \\\hline
SC & N & poor quality and heavily outdated content. video quizzes were broken more often than not. module quizzes questions and answers were vague and poorly constructed resulting in choosing incorrect answers.reading material was of poor quality with many "industry professionals" not being professional enough to perform a simple spelling and grammar check. \\ \bottomrule
\end{tabular}
\caption{Reviews samples from CR23k}
\label{tab:crsample}
\end{table}

\begin{figure}[ht]
    \centering
    \begin{subfigure}[b]{.49\textwidth}
        \centering
        \includegraphics[width=\textwidth]{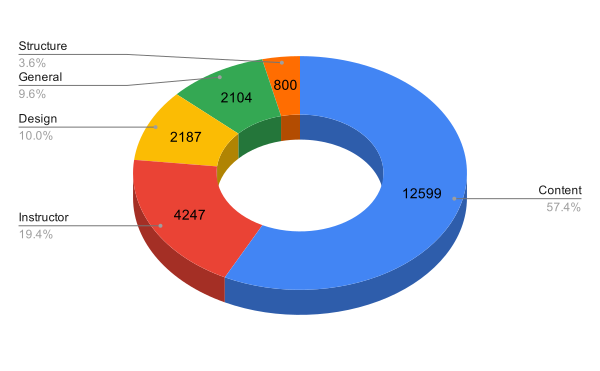}
        \caption{Course label distribution}
        \label{srfig1:a}
    \end{subfigure}
    \hfill
    \begin{subfigure}[b]{.49\textwidth}
        \centering
        \includegraphics[width=\textwidth]{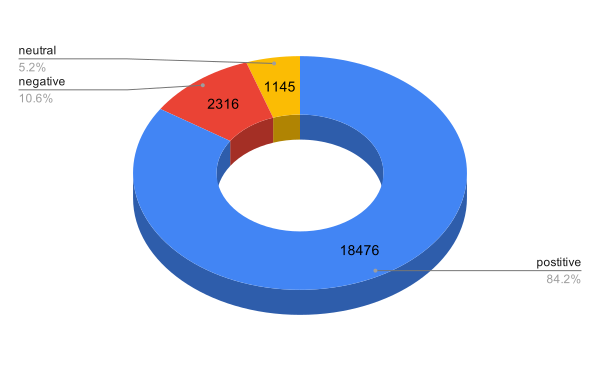}
        \caption{Sentimental polarity distribution}
        \label{srfig1:b}
    \end{subfigure}
    \caption{Detailed information of CR23k Dataset}
    \label{fig:cr_label}
\end{figure}

This dataset is mainly in English and manually labelled with three sentimental polarities: positive, negative, and neutral. The percent rate of each sentimental polarity in the whole dataset is not equal, in which the positive reviews count most, having 18476 records while negative reviews and neutral reviews count 2316 and 1145, respectively. Also, five aspects of each review are labelled containing Content, Instructor, Design, General, Structure. The detailed information of this dataset has been shown in \ref{fig:cr_label}.

\subsection{CR100k}
\begin{figure}[hbp!]
    \centering
    \begin{subfigure}[b]{.49\textwidth}
        \centering
        \includegraphics[width=\textwidth]{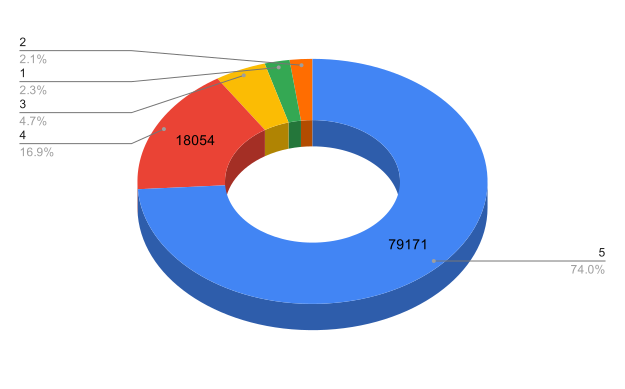}
        \caption{Rating score distribution}
        \label{srfig:a}
    \end{subfigure}
    \hfill
    \begin{subfigure}[b]{.49\textwidth}
        \centering
        \includegraphics[width=\textwidth]{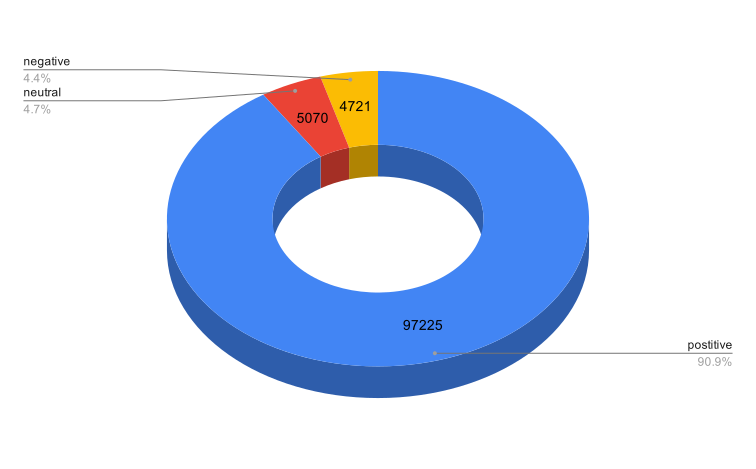}
        \caption{Sentimental polarity distribution}
        \label{srfig:b}
    \end{subfigure}
    \caption{CR100k Dataset with rating scores from 1 to 5}
    \label{fig:sr_label}
\end{figure}

The second dataset includes 107016 reviews obtained from Kaggle \footnote{https://www.kaggle.com/septa97/100k-courseras-course-reviews-dataset}. The author scraped reviews of different courses from the online learning platform -- Coursera and published it on Kaggel. The dataset also includes rating scores each user gave to the course when they wrote the review. The rating value is from 1 to 5, and the rating with five counts mostly having 79171 records. The reviews samples of each rating score have been displayed in Table \ref{tab:100k+sample}. The more detailed presentation of each label has been displayed in Figure \ref{fig:sr_label}. This dataset is not manually labelled with the same sentimental polarity for each review. Instead, rating information is utilized to label those reviews. In order to simplify the labelling process, we classify reviews with rating 4 and 5 as positive polarity, reviews with rating 3 as neutral polarity, and rating less than three as negative polarity. According to our common sense, this labelling method would work for most reviews, and a random check on each category was executed to see whether this labelling method works fine. According to our randomly check results, most reviews fall into the right category. We think they would increase the robustness of the models we would train later for those in the wrong category, so we do not remove them. The detailed distribution of each sentimental category has been shown in Figure \ref{fig:sr_label}. Hereafter, we would use CR100k to refer to this dataset.

\begin{table}[hbp!]
\centering
\begin{tabular}{@{}lp{0.85\textwidth}@{}}
\toprule
\textbf{Label} & \textbf{Review} \\ \midrule
5&This class is very helpful to me. Currently, I'm still learning this class which makes up a lot of basic music knowledge. \\
4&Really nice teacher!I could got the point eazliy\\
3&Good content, but the course setting does (at least for me) not allow learn the content long term due to missing reading material. \\
2&This course does not say anything about digitization which is the core subject of the digital wave. \\
1&A lot of speaking without any sense. Skip it at all cost\\
 \bottomrule
\end{tabular}
\caption{Couseara reviews samples with rating score for CR100k dataset}
\label{tab:100k+sample}
\end{table}

\begin{table}[hbp!]
\centering
\begin{tabular}{@{}lp{0.75\textwidth}@{}}
\toprule
\textbf{Class} & \textbf{Review} \\ \midrule
Game&this was an okay game but not very fun . a bit confusing at times because words didn't match up so i deleted it .\\
Game&i really enjoy this game i can forget about the real world . i can use my imagination and have fun . i would recommend this game for all . \\
App&i can't believe how much my 5 year old loves this app . it is really cute , has a ton of levels , and is free ! i am very surprised it was free !\\
App&this app is a waste of time . very bad movies . not much of a selection . bad not for me . sorry\\
 \bottomrule
\end{tabular}
\caption{Samples for dataset AMR}
\label{tab:amazonsample}
\end{table}

\subsection{AMR}

\begin{figure}[hbp!]
    \centering
    \includegraphics[width=0.7\textwidth]{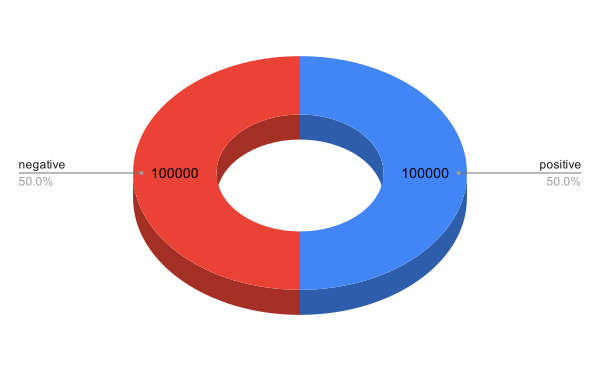}
    \caption{Amazon game and application reviews}
    \label{fig:amazon senti}
\end{figure}

Amazon game and movie reviews is the last dataset used in our project. This dataset is used and published in the paper \cite{liu2020catgan}, which contains 200000 reviews of both movie and game from Amazon. The sample of reviews of the first and last ten reviews have been displayed in table \ref{tab:amazonsample}. Each review in this dataset is classified into only two categories, either positive or negative. It is a balanced dataset in which the number of positive and negative reviews are the same that is 100000. 
The detailed information of these three datasets has been shown in Fig. \ref{fig:amazon senti}. Later in the thesis, we would use AMR to refer this dataset.

\section{Results}

We can see that models trained with a balanced dataset generally have better performance both on accuracy and F1-score than models trained with the original imbalanced dataset. The results demonstrate that balancing the highly imbalanced datasets using the SOTA text generation GAN model can improve the performance both on accuracy and F1 score of sentiment classification models. Also, the results show that the more unbalanced the dataset, the more significant the performance improvement of the sentiment classification models enhanced using the method proposed in this study.

Suppose we compare results of one dataset, for example, CR23k, we see that the more complicated models are easier to be influenced by imbalanced dataset. This may be because the models with complex network structure usually have more parameters than the usual model. They may need more data to be fed in order to achieve outstanding performance. Therefore, when the dataset is not enough or imbalanced in certain classes, those models are more likely to be influenced by insufficient or unbalanced datasets. The results of machine learning algorithms and RNN also indicate the same. The network structure of machine learning algorithms and RNN is relatively simple compared to other deep learning models. Hence, performance improvement before and after balancing the dataset is not very apparent for the CR23k dataset, whose ratio between positive and negative classes is not huge. Also, we can see the most considerable improvement after balancing the dataset is from LSTM and GRU model with BERT transformers. This can also be explained by the fact that BERT transformer used in the experiment contains 11 BERT layers that are composed of attention, self-output, intermediate, and output layers. When the dataset is imbalanced, LSTM and GRU models with BERT are more likely to focus more on the category with the most significant number (that is, positive class in our experiment) but ignore the other class with fewer data in the whole dataset. Therefore, when we balanced our dataset and use this dataset to re-train the models with BERT transformer, they have generally higher accuracy and F1-score. The results of different models for dataset CR100k also demonstrates the same as CR23k.

If we compare models for different datasets, we can see that when the gap between positive and negative or neutral classes widens in that dataset, the performance improvement of sentiment classification models is greater. The ratio between positive and neutral category is $18746\div 2316 \approx 8.094 $ for dataset CR23k and $74191\div 2602 \approx 28.51$ for dataset CR100k. The average improvement of accuracy of all tested models, including machine learning algorithms, is $2.039\%$ for CR23k dataset and $4.822\%$ for CR100k dataset. The improvement for CR100k dataset is more than two times the average improvement for CR23k dataset.

In all, when we come back to our problem statement that the impact of synthetic text generation on sentiment classification task of highly imbalanced dataset, we can conclude that after balancing the highly imbalanced dataset the highly imbalanced training dataset using CatGAN text generation model, the performance on accuracy and F1-score is improved that accuracy increases 2.039\% and 4.822\% for CR23k and CR100k dataset, and F1-score increases 2.79\% and 9.208\% for CR23k and CR100k dataset. Results also indicate that the increasing degree for CR100k is always higher than CR23k. The average increasing degree for deep learning is higher than machine learning algorithms, and the average increasing degree for more complex deep learning models is higher than simpler deep learning models experiments.
\begin{table}[ht]
\centering
\begin{tabular}{|c|c|l|r|r|}
\hline
\multicolumn{3}{|l|}{}                                                                           & \textbf{CR23k} & \textbf{CR100k} \\ \hline
\multirow{6}{*}{Difference} & \multirow{3}{*}{\textbf{Accuracy (\%)}} & Deep learning    & 3.09          & 6.64           \\ \cline{3-5} 
                                    &                                         & Machine Learning & 0.46         & 2.10          \\ \cline{3-5} 
                                    &                                         & Overall average  & 2.04          & 4.82           \\ \cline{2-5} 
                                    & \multirow{3}{*}{\textbf{F1-score (\%)}}      & Deep learning    & 6.06        & 11.30          \\ \cline{3-5} 
                                    &                                         & Machine Learning & -2.12       & 6.08         \\ \cline{3-5} 
                                    &                                         & Overall average  & 2.79        & 9.21         \\ \hline
\end{tabular}
\caption{Summary statistics of the degree of difference of sentiment classification models for different data sets after balancing the dataset}
\label{tab:dis}
\end{table}

\section{Conclusion}
\label{ch:con}
Schools and universities have switched to online teaching from on-campus teaching due to the COVID-19 pandemic, and mining students' reviews towards online courses become critical in helping teachers and schools understand students' feedback and need as well as improving online teaching quality. But dataset imbalance is a quite often problem for sentiment classification within the education domain, which means there are much fewer neutral and negative reviews than positive reviews. The highly imbalanced dataset problem would influence the performance of sentiment classification models. We aimed to use SOTA GAN models to synthesize text, and analyze the impact of synthetic text generation on the sentiment classification task of the highly imbalanced dataset using deep learning and machine learning.

 Two SOTA category-aware GAN models are trained with the imbalanced dataset. Both GAN models are trained with 250 epochs. We compared metrics results and generated samples of these two samples on three different datasets mentioned above. Finally, the category-aware GAN model with a hierarchical evolutionary algorithm that is able to generate higher-quality text without losing text diversity compared with SentiGAN is selected to generate text to balance the highly imbalanced training dataset for sentiment classification.

The imbalanced and synthetic balanced datasets are obtained from the last experiment step. Same machine learning algorithms and deep learning models are trained on synthetic balanced and imbalanced dataset from the different dataset, respectively. The results indicate that compared with the original imbalanced dataset, the performance on accuracy and F1-score of the model trained on synthetic balanced dataset from CatGAN text generation model, is improved. Specifically, accuracy is increased from 2.039\% to 4.822\% for CR23k and CR100k dataset, whereas F1-score is increased from 2.79\% to 9.208\% for CR23k and CR100k dataset. Also, the results show that the improvement for CR100k is higher than CR23k. Also, the average performance improvement for deep learning is higher than machine learning algorithms.

Due to time limitation, we have not extended our experiments on more complex sentiment analysis deep learning models such as aspect-based sentiment analysis model to see how those more sophisticated models would behave on the synthetic balanced dataset. Nevertheless, these four models are the necessary parts for most NLP deep learning models used for sentiment analysis. So we infer that the performance improvement of these four models would more or less improve the performance of models with more complex architectures. Besides, just GAN text generation models are exploited while some newest transformer-based text generation model such as GPT-3 has not been tested yet, and the experiments are limited within the education domain. In the future, researchers could exploit different type of text generation and more complex sentiment analysis models in order to have a complete picture of the impact of synthetic text generation on the sentiment classification task of the highly imbalanced dataset. Besides, researchers can also try to construct a new sentiment analysis model that can avoid the influence of a highly imbalanced dataset.

\bibliographystyle{unsrt}
\bibliography{access.bib}

\end{document}